# Distributed learning optimisation of Cox models can leak patient data: Risks and solutions

Authors:

Carsten Brink[1,2]; Christian Rønn Hansen[1,2]; Matthew Field[3,4]; Gareth Price[5]; David Thwaites[6] ; Nis Sarup[1]; Uffe Bernchou[1,2]; Lois Holloway[3,4,6,7]

1) Laboratory of Radiation Physics, Department of Oncology, Odense University Hospital, Odense, Denmark
2) Department of Clinical Research, University of Southern Denmark, Odense, Denmark
3) South Western Sydney Clinical School, Faculty of Medicine, UNSW, Sydney, New South Wales, Australia
4) Ingham Institute for Applied Medical Research, Liverpool, New South Wales, Australia
5) The University of Manchester, Manchester Academic Health Science Centre, The Christie NHS Foundation Trust, Manchester, UK
6) Institute of Medical Physics, School of Physics, University of Sydney, Sydney, New South Wales, Australia
7) Liverpool and Macarthur Cancer Therapy Centres, Liverpool, New South Wales, Australia

e-mails:

**Carsten Brink Carsten.Brink@rsyd.dk; Christian Rønn Hansen christian.roenn@rsyd.dk**

**Matthew Field matthew.field@unsw.edu.au; Gareth Price Gareth.Price@manchester.ac.uk**

**David Thwaites David.Thwaites@sydney.edu.au; Nis Sarup Nis.Sarup@rsyd.dk**

**Uffe Bernchou Uffe.Bernchou@rsyd.dk; Lois Holloway Lois.Holloway@health.nsw.gov.au**

**Abstract**

Medical data are often highly sensitive, and frequently there are missing data. Due to the data's sensitive nature, there is an interest in creating modelling methods where the data are kept in each local centre to preserve their privacy, but yet the model can be trained on and learn from

 data across multiple centres. Such an approach might be distributed machine learning (federated learning, collaborative learning) in which a model is iteratively calculated based on aggregated local model information from each centre. However, even though no specific data are leaving the centre, there is a potential risk that the exchanged information is sufficient to reconstruct all or part of the patient data, which would hamper the safety-protecting rationale idea of distributed learning.

This paper demonstrates that the optimisation of a Cox survival model can lead to patient data leakage. Following this, we suggest a way to optimise and validate a Cox model that avoids these problems in a secure way. The feasibility of the suggested method is demonstrated in a provided Matlab code that also includes methods for handling missing data.

**Introduction**

Digital health initiatives promise that analyses of routinely collected 'real-world' patient data can provide insight with which we can improve clinical care. However, working with medical data can be challenging: 1) the data often contain sensitive information and require adequate protection and privacy; 2) there are often missing data; and 3) within a given centre, the amount of data is often too limited for robust model creation.

The limited amount of patient data can be addressed by pooling data from different centres in larger national or international databases. This approach has been used extensively and is an efficient approach. However, in many cases, it is difficult to share the data due to its sensitivity and associated legal and governance issues, particularly if data needs to be shared across national and/or jurisdictional borders. As such, there is increasing interest in the distributed learning (federated learning, collaborative learning) approach in which sensitive data remain stored locally in each centre to preserve privacy, but a model is optimised iteratively across multiple centres where each local centre only provides aggregated information from its own data [1-4]. One such method could be the optimisation of a likelihood function in which the individual part of the likelihood can be calculated locally and combined in a central server.

Distributed learning networks have previously shown their ability to create models without pooling the data in a central database [3]. There has also been work on how to optimise a Cox survival model in a distributed learning network [5]. However, even though the distributed learning system does not directly exchange specific patient data, there is a risk that the optimisation of a Cox model can leak patient information. Such leakage would hamper the entire safety-protecting rationale of distributed learning.

This paper demonstrates a framework supporting multiple data imputations that can be used to optimise a Cox model using distributed learning without patient data leakage. From this, the paper demonstrates methods for possible model validation without leaking patient data. The code for model optimisation is written in Matlab and this is made available as open-source. To ensure the code is kept easy to read, run and understand, the actual exchange of messages between the involved computers is simplified such that the code can be executed on a stand-alone computer. The supplied code is currently used within the AusCAT distributed learning network in a collaboration between Odense University Hospital (Denmark), Liverpool Hospital (Australia), and Christie Hospital (UK). The details of the network infrastructure and communication method have been described separately [6,7]. The current article aims to provide a method of performing and validating Cox distributed learning without the risk of patient data leakage. The suggested method also integrates data imputation since full information is rare within medical datasets.

**The Cox model and data leakage**

The hazard for the individual patient within the Cox proportional hazard model is defined as:

$$\lambda(t|\bar{X}_i) = \lambda_o(t) \exp(\bar{X}_i \bar{\beta}) \qquad Eq.\ 1$$

where $\bar{X}_i$ is a row vector with clinical measurements for the $i$'th patient, $\bar{\beta}$ is the related column vector of regression constants for the entire cohort and $\lambda_o(t)$ is a non-parametrised time-dependent hazard function related to the hazard for a fictive patient with all values in $\bar{X}_i$ being equal to zero. The related survival function will be:

$$S(t) = e^{-\int_0^t \lambda(t'|\bar{X}_i)dt'} = e^{-\int_0^t \lambda_0(t')dt' \exp(\bar{X}_i\bar{\beta})} = e^{-\Lambda_0(t)\exp(\bar{X}_i\bar{\beta})} \qquad Eq.\ 2$$
$$= S_0(t)^{\exp(\bar{X}_i\bar{\beta})}$$

where $\Lambda_0(t)$ is the integrated baseline hazard, and $S_0(t)$ is the baseline survival curve. The survival times of the patients are given as a pair of values $(Y_i, C_i)$ where $Y_i$ is the observed survival time for the $i$'th patient and $C_i$ is an indicator variable being one for events and zero for censoring. If all values of $Y_i$ are unique, and $Y_i$ is related to an event ($C_i = 1$) the partial-likelihood of the $i$'th patient can be written as (see Appendix 1 for the definition, including ties as well as derivatives of the likelihood):

$$L_i(\bar{\beta}) = \frac{\lambda(Y_i|\bar{X}_i)}{\sum_{j:Y_j \geq Y_i} \lambda(Y_i|\bar{X}_j)} = \frac{\lambda_o(Y_i)\exp(\bar{X}_i\bar{\beta})}{\sum_{j:Y_j \geq Y_i} \lambda_o(Y_i)\exp(\bar{X}_j\bar{\beta})} = \frac{\exp(\bar{X}_i\bar{\beta})}{\sum_{j:Y_j \geq Y_i} \exp(\bar{X}_j\bar{\beta})} \qquad Eq.\ 3$$
$$= \frac{\theta_i}{\sum_{j:Y_j \geq Y_i} \theta_j}$$

where $\theta_i = \exp(\bar{X}_i\bar{\beta})$ and the index on the summations reads all the $j$'th values for which $Y_j \geq Y_i$, so the sum is over all patients at risk at the time $Y_i$, which turns out to be problematic in a distributed learning context, as discussed below. The partial likelihood function over all patients is calculated as a product of likelihoods for all event times (censoring times only indirectly adds information). The logarithm of the partial-likelihood can therefore be written as:

$$l(\bar{\beta}) = \sum_{i:C_i=1} \log\left(L_i(\bar{\beta})\right) = \sum_{i:C_i=1}\left(\bar{X}_i\bar{\beta} - \log\left(\sum_{j:Y_j \geq Y_i} \theta_j\right)\right) \quad \text{Eq. 4}$$

The likelihood that should be optimised is a simple sum over all event-times. However, within the summation, there is a need to know

$$\sum_{j:Y_j \geq Y_i} \theta_j \quad \text{Eq. 5}$$

In distributed learning each institution must share this summation (eq 5) for each local event time ($Y_j$) in order to allow the complete partial log-likelihood (eq 4) to be compiled centrally. This can lead to information leakage. The leakage problem is straightforward to understand in the simple case of unique event times (no ties), and no censoring. In that case, for the patient surviving the longest e.g. patient $z$, each institution would need to share:

$$\theta_z = \exp(\bar{X}_z\bar{\beta}) \quad \text{Eq. 6}$$

since the sum would only contain one term. Likelihood optimisation across the centres will typically be performed iteratively; therefore, a set of $\theta_z$ for different values of $\bar{\beta}$ will be shared between the centres. If the number of iterations is larger than the number of patient variables, there will be sufficient information to reconstruct $\bar{X}_z$ outside the local centre. After determination of $\bar{X}_z$, it is possible to determine the clinical values of the patient with the next longest survival and recursively determine all the clinical values for all the patients. Thus such an approach would leak the entire patient information from the local centres. If the data did contain ties and censoring, the reconstruction is more challenging, but it will, in many cases, still be possible to reconstruct at least part of the clinical data.

**Alternatives to the standard Cox proportional hazard model**

The leakage problem described above is related to the use of a non-parametric hazard function $\lambda_o(t)$, which in turn leads to the partial likelihood function. If a parametrised version was used for $\lambda_o(t)$, the partial likelihood shown in eq. (3-4) would be replaced with a full likelihood function, which could be calculated for each patient individually without knowing the outcome of the other patients. Thus, in that case, there would be no need to share the information stated in equation (5); only the parameters related to the parametrisation of the baseline hazard and the regression constants need to be shared. The baseline hazard has often been parametrised by assuming that the baseline event distribution $(-dS_0(t)/dt = S_0(t)\lambda_o(t))$ is a differential Weibull distribution $(bkt^{k-1}e^{-bt^k})$ (see e.g. [8, 9]), which leads to $\lambda_o(t)$ being equal to $kt^{k-1}$, where k is a shape parameter and b is a scaling parameter (for $k$ =1, this results in a constant event rate and exponential survival functions; so the parametrisation provides quite a flexible class of functions). Thus, a parametrisation of the baseline function is a possible solution to the data leakage problem discussed above.

Even though a parametrisation of the baseline hazard is a possible solution to the data leakage problem, in this paper we recommend another approach. The reason to use a different approach is that fully parametrised hazard functions are much less used, within medical statistics, relative to the non-parametrised Cox model. We propose using the non-parametrised Cox proportional hazard model, described above, but stratified per centre. The stratification introduces centre-specific individual baseline hazards ($\lambda_o(t)$); thus, the summation in the denominators of equations (3) will only be performed over the patients within each individual centre, which removes the need to share these denominators between the institutions. The likelihood described in equation (3) can thus be calculated locally, and the product of the local likelihoods will be the likelihood that needs to be optimised during the distributed learning. The stratification introduces additional flexibility within the model (individual local baseline hazards), which might not be optimal, e.g. if many centres with very few patients are included in the study. However, on the other hand, the centre-specific baselines introduced due to the centre stratification can help identify cohort differences between institutions beyond that described by the variables explicitly included in the model. In many medical research studies, such centre differences are likely, particularly if centres from different countries are included. Thus, in summary, the frequently used Cox model can be used without data leakage if stratification is performed per centre. Furthermore, the stratification can help to identify cohort differences that otherwise might not be identified.

**Parameter selection and uncertainty confidence intervals**

Within medical research, parameter selection for a Cox model is often performed by stepwise backward or forward parameter selection processes based on p values within each selection step. This approach is easy to use but is statistically unsound, e.g. due to problems related to co-linearity and because of the large number of calculated p values that in general leads to an overestimation of the $\bar{\beta}$ values (e.g. see page 68 of Harrell [10]). Therefore, the current paper suggests using best subset selection based on cross-validated likelihoods for parameter selection. This approach is only possible as long as the number of regression parameters is not too large (e.g. below 20), since otherwise, best subset selection will be too time-consuming to perform. In the current paper, bootstrap sampling has been chosen as the method to determine both the estimate of the $\bar{\beta}$ values, but also the related confidence interval. The best subset method could select the model that contains the combination of initially proposed variables that results in the best cross-validated likelihood. However, such an approach might choose a more complex model that is only slightly better than another model containing fewer parameters. Thus, to reach a simpler model with similar performance, it can be advisable to use the "one standard-error rule" suggested by James et al. [11] (page 214). They recommend selecting the simplest model where the cross-validated likelihood is within one standard deviation of the best model.

Bootstrap sampling is performed at each institution. Thus each model is optimised a certain number of times, and within each optimisation, a sample of the same size as the original data are drawn from the local patient data with replacement. The median value per component of the distribution of obtained $\bar{\beta}$ values will be reported as the $\bar{\beta}$ estimate, and the distribution will also be used to estimate the two-tailed confidence intervals, defined as the range between the percentiles $\alpha/2$ and $1 - \alpha/2$ for the $(1 - \alpha)$-confidence interval, where $\alpha$ is pre-defined at the optimisation computer e.g. as 5%. Furthermore, the bootstrap method can also be used for cross-validation, as described by James et al. [11]. For each bootstrap, the model is optimised based on the patients within the sample (in-bag patients – some patients will be present more than once) but can then be cross-validated on the patients not selected for the given bootstrap (out-of-bag patients). For samples larger than 10 patients, the probability that a given patient is not included in a specific bootstrap is in the range of 35-37% (the limiting value for infinite samples are $e^{-1} \approx 37\%$) thus, approximately one-third of the patients will be out-of-bag patients and can be used to calculate the cross-validated likelihood for each bootstrap. As described above, the cross-validated likelihoods are used to select the final model. However, a calculated likelihood depends on the number of patients included in the calculation. The number of out-of-bag patients will fluctuate due to the statistical nature of the sampling process. Such a fluctuation of the cross-validated values is obviously not warranted; thus, the calculated cross-validated values need to be adjusted depending on the size of the out-of-bag patient group. As seen in equation (4), the log-likelihood is a sum over all event times within a given institution; thus, one might initially think that the cross-validated values scale linearly with the number of events in the out-of-bag patients. However, it has been shown that the log-likelihood is proportional to the number of patients and not the number of events [12]. Thus, to avoid the unwanted fluctuation of the cross-validated log-likelihood, the obtained value is adjusted by dividing by the number of out-of-bag patients and multiplying by the number of patients within the local cohort.

**Missing data**

In many medical datasets, there is some missing data. A potential solution to this problem is to only include the patients for which the complete information is available, but this could lead to a strong bias and dismissal of a large part of the available data. An alternative is to use data imputation, where a simple approach would be to estimate the missing values by the mean or mode of the available values of a given variable. However, using a single imputed value is somewhat counterintuitive since it indicates that it is possible to predict all the missing values based on the other variables. If that indeed were the case, the values would not really be missing since they could be estimated from the available information. An alternative is to use multiple imputations in which imputation is done multiple times to reflect the imputed values' uncertainty. Thus, multiple imputations are often preferred, which fits nicely into the chosen bootstrap approach in this paper. Multiple imputations are performed using the ideas of Multivariate Imputation by Chained Equations (MICE) as implemented in the statistical package R by van Buuren [13, 14], but here re-implemented in Matlab (the implementation is mainly based on algorithm 3.1 to 3.5 in [14]). The approach makes a prediction model for each variable based on the other variables and uses the model to estimate the missing data values. The idea of multiple imputations is to make multiple datasets that contain the same original measured values but have different imputed values for the missing information. The intended analysis is then performed on all the new datasets and the results are combined to a single result. If there is more than one missing value for any patient, the imputation of a given variable can not be based on all the other variables due to missing values within the predictors. In that case, an initial mean/mode imputation will be performed, and the imputation process will be performed iteratively to stabilise the predicted values. The number of needed multiple imputations depends on the degree of missing values, but typically less than 20 imputations are needed, and often the numbers are as low as 5 to 10 imputations [15]. In the current setup, the number of imputations is equal to the number of bootstraps, which can be held at a minimum value of 20 for model selection and will be much larger for the model evaluation. The current setup, therefore, ensures sufficient number of imputations for a stable prediction result – and in many situations, the number of imputations could without loss of predictability be reduced, but it is simply convenient to create as many imputations as bootstraps. It might be important to stress that the imputation process for all multiple imputations only estimates the missing values but keeps all the measured data without any changes.

In the current work, bootstrapping is used to estimate the uncertainty of $\bar{\beta}$ as well as a method for model selection (see above). It thus seems natural to produce as many imputed datasets as the requested number of bootstraps. Therefore for each set of data created by imputation (identical to original data except for the imputation), one bootstrap is carried out, and the calculation of likelihood is performed within that bootstrapped dataset. This method will produce a distribution of $\bar{\beta}$ values and the final reported values will be the median value of each variable within $\bar{\beta}$. The entire process is outlined in figure 1, and more detailed explanation is provided in the following section.

**Calculation process overview**

Figure 1 shows an overview of the calculation process, exemplified for two contributing centres. Part A is related to model selection, while part B is for the model performance. For the model selection part, the diagram represents the flow for one potential model candidate within the best subset selection approach (e.g. one of the 2047 models for 11 predictors as in the simulated data). In the top box, the optimisation computer requests each institution to make n bootstraps using an initial beta value of zero for all bootstraps. Based on that instruction, the local computers create n bootstrapped versions of the original data, but before each bootstrap, a data imputation is performed; thus, an independent imputation is performed for each bootstrap. The likelihood (and related gradient and Hessian and cross-validated likelihood) is calculated for each centre and bootstrap. All these values are forwarded to the optimisation computer. In the optimisation computer, the values from the different centres are added together for each bootstrap separately – resulting in n values each summed over the institutions. Based on these values, new beta values (one beta vector for each bootstrap) is calculated and returned to the local computers. The local computer then evaluates the updated beta values, and the process continues until convergence. Finally, all the beta values, likelihoods, and cross-validated likelihoods are collected in the optimisation computer. The cross-validated likelihoods for all the model candidates within the best subset selection process are used to select the final model (described in the parameter selection section above). The performance evaluation of the chosen model is seen in part B of figure 1. Initially, each local computer is instructed to make m bootstrapped datasets with imputation as in part A of the figure (m likely much larger than n used for the model selection, in the simulated data n=20, m=1000). The selected model is then evaluated as in part A of the figure, resulting in m sets of beta values. The model performance (in the figure exemplified as the Kaplan-Meier curve) is calculated for each center and bootstrap based on the bootstrap-related beta value or the median beta over all datasets (see the section model performance for details). The model performances at the local centers are obtained by calculating the median value of the performance of the individual bootstraps at the local center. Similarly, the local confidence intervals are computed using pre-defined percentiles, e.g. reflecting the 95% confidence interval. For performance values that are functions (e.g. the Kaplan Meier curve), the data are locally aggregated over a minimum number of patients pre-defined at the local computer. All the performance measures are finally reported to the optimisation computer.

**Server communication**

The specific method for communication between the centres has been presented by Field et al. [6] and hence will not be described in detail here. Briefly, the system consists of a central optimisation computer, a computer hosting an internet service, and one local computer at each participating institution. The optimisation and local computers are not directly connected but do all contact the internet service at regular intervals. The computers can push and fetch messages to and from the internet service, which is a set of JAVA applications running in Tomcat [ref https://tomcat.apache.org/]. The optimisation computer can put a message on the internet service asking the local computers to calculate, e.g. the partial likelihood for a given set of $\bar{\beta}$ values. The next time the local computers contact the internet service, they will fetch the instruction and start performing the calculation based on the local patient data. When the optimisation computer detects that all the local computers have reported the request information (e.g. likelihood, gradient, and Hessian matrix) to the internet service, the information will be fetched and combined. If the log-likelihood is calculated as a sum of the local values, all patients will have the same weight within the optimisation which is the approach used in the simulations in this paper. In case of unequal patient numbers between centres, the centre with the most patients will be weighted more than those with few patients. In some situations, it might be preferred that all centres have the same weight independent of the number of provided patients. In that situation, the local log-likelihood needs to be adjusted to reflect a specific number of patients before adding them together. In the supplied simulation data, the local centres are returning, for each $\bar{\beta}$ iteration, the local log-likelihood and the related gradient and Hessian as well as the cross-validated log-likelihood. The local values are summed in the optimisation computer and used to determine the $\bar{\beta}$ value of the next iteration ($\Delta\bar{\beta} = -\bar{\bar{H}}^{-1}\bar{G}$ for $\bar{\bar{H}}$ and $\bar{G}$ being the current Hessian and gradient of the total log-likelihood, respectively), and the local computers are then instructed to repeat the calculations on updated $\bar{\beta}$ values. This process will then continue until convergence. After convergence, the communication computer can request information about the final model's performance on the local computers (see section "Demonstration of model performance" below).

**Included software and configuration**

The required communication between the distributed network computers depends on the given task, e.g. in the current case, stratified Cox modelling. Therefore, task-specific programs are needed for the optimisation computer and the local computers. The programs for the local computers can be distributed from the optimisation computer through the internet service. A Matlab version of these programs for the stratified Cox is included as part of this paper (https://osf.io/vuqrc/?view_only=86eb4293390f4eefbb4b21fcbe4699d8). For clarity, the internet communication part is not included. The local computers are simulated as Matlab classes (see detailed description in appendix). The delivered code should be executable after download and it also includes the demonstration data used for the current work. We hope that making the program publicly available will make it possible for users to combine the code directly with their choice of a specific communication system (see appendix 2 for a description of the program).

**Simulated data**

The distributed Cox analysis described above has been demonstrated based on simulated data. To make the simulation clinically relevant, the simulated data reflects the model of survival of Head and Neck cancer patients after radiotherapy by Egelmeer et al. [16]. Their model for survival after 2 and 5 years is presented as a nomogram in figure 1 of their paper. By digitising this, the regression constants have been extracted, and the linear predictor ($\bar{X}_i \bar{\beta}$ of equation 1) for each patient can be calculated based on the nomogram, as:

$$\begin{aligned}\bar{X}_i\bar{\beta} = &\ 0.04136 \times Age_i - 0.400 \times hemoglobin_i - 0.03506 \times eqd2t_i \\ &+ 0.1943 \times T2_i + 0.7965 \times T3_i + 1.4546 \times T4_i \\ &+ 0.3764 \times Nplus_i + 0.8353 \times genderMale_i + 0.2695 \times NonGlottis_i\end{aligned} \quad Eq.\ 7$$

where $Age$ is the patient age in years at the start of radiotherapy, $hemoglobin$ is the hemoglobin level in units of mmol/L before radiotherapy, $eqd2t$ is the prescribed radiotherapy dose in 2 Gy equivalent doses and corrected for the time duration of the radiotherapy treatment [16, 17], $T1$-$T4$ are dummy variables indicating the clinical stage of the tumour with T1 being the reference level, $Nplus$ is an indicator variable indicating nodal cancer involvement, $genderMale$ is indicator variable for the male gender, and $NonGlottis$ is an indicator of cancer outside the glottis.

All the predictors of equation 7 were included in the simulation, and their values were randomly sampled based on a clinical cohort. The survival time for the individual patients was sampled from a Cox distribution using the linear predictor of equation 7. Within each centre, a different but constant baseline hazard ($\lambda_o$ of equation 1) was used. Data values beyond 60 months were censored. To simulate lost to follow-up censoring during the observation time a second simulation was introduced with a slightly lower baseline hazard. If the second simulation resulted in a survival time that was less than the original simulated value, the original survival time would be censored at the time point of the second simulation. Besides the predictors in equation 7, four additional randomly generated variables were included as potential predictors i.e. two numerical and two categorical variables. The numerical variables, named Cont 1 and 2, were simulated using a standard normal distribution, while the two categorical, named Factor1 and 2, were simulated as binary variables with 50% probability of each level. These additional variables were added to demonstrate the variable selection mechanism.

The simulated data included 1000 patients within each centre. 20% of the simulated predictors of equation 7 were randomly set to be missing. It might be preferable to include only patients with a maximum number of missing variables in some situations. This possibility is included in the code as an option, and the data below are shown for patients with a maximum of one missing variable, resulting in 569, 558, and 563 patients used for analysis within the individual centres. A simulation using all patients independent of the number of missing values and a simulation using only the patients with no missing values have been made, and both simulations are selecting the same predictors for the final model (data not shown).

**Demonstration of model selection**

The variable selection process included all the seven variables from equation 7 (counting the T levels as one variable) and four additional random variables as described above, i.e. a total of 11 variables. The best subset selection process described above involves all possible subsets of the variables. For 11 variables this is 2048 (2^11) if the null model is included. In the simulation code, the likelihood of the null model is subtracted from all the other models; thus, the null model is not analysed separately. The negative cross-validated log-likelihoods for all 2047 models are shown in figure 2 as a function of the number of variables within the model. The error bars shown in the figure are the standard deviation of the 20 bootstraps used during the parameter selection process. The best performing model is a model including 7 parameters. Not too surprisingly, the 7 parameters in the best performing model are those of equation 7, which was used to simulate the survival times. In figure 2 it can be seen that all models based on four parameters are outside the range of one standard deviation, while the best performing model including five parameters is within the one-standard-error rule. Thus, the selected model includes five parameters. The five parameters in the selected model are the same as those for the best performing model, except that $Nplus$ and $NonGlottis$ are not included. Both $Nplus$ and $NonGlottis$ were both included during the creation of the simulated data and should therefore be significant for an infinitely large patient population. However, the omission of $Nplus$ and $NonGlottis$ simply indicates that their contribution to the overall likelihood is too small to be identified in the current cohort. The small likelihood impact is reflected by the small difference in the cross-validated likelihood between the 7 and 5 parameter models seen in figure 2.

**Demonstration of model performance**

Based on the five parameter model selected above, the model was bootstrapped 1000 times to have sufficient information to calculate a stable median and confidence interval for all regression parameters. The distribution of the bootstrapped regression parameters can be seen in figure 3. The value reported as the estimator for each regression constant will be the median bootstrap value (indicated by the vertical red line in figure 3), and the confidence interval will be defined by the range of the 95% most central values (range between the 2.5% and 97.5% percentiles), which is shown as red dashed lines in figure 3.

Besides the regression parameters, a total specification of a stratified Cox model involves the specification of the baseline hazard, which might be different at the individual centres. The baseline hazard can be calculated using the Breslow estimator (page 485 of [10]):

$$\widehat{\Lambda}_0(t) = \sum_{i:Y_i<t} \frac{m_i}{\sum_{j:Y_j \geq Y_i} \exp(\bar{X}_j \bar{\beta})} \qquad Eq.\ 8$$

Where $m_i$ is the number of events at time $Y_i$, thus $m_i$ is zero for censoring times and can be larger than one in case of tied events. To further prevent information leakage, the cumulative hazard stated in equation 8 will only be reported at time points for which a fixed number of event times occur between individual reported time points. The number of aggregated patients between time points can be defined in the config file at the local centre. The reported information from each individual centre is only the median of all bootstraps per time point as well as a pre-defined confidence interval. Thus only one version of $\widehat{\Lambda}_0(t)$ is reported to the optimisation computer, and it is even in an aggregated version; thus, it is not possible to deduce the patient sensitive information from the baseline information. In the simulated data, all model data provided to the optimisation computer is aggregated over five patients, which is a number defined at each local centre. The baselines obtained from the simulated data are shown in figure 4 together with the 95% confidence interval. The confidence intervals are based on the variation within the different bootstraps at the individual centres.

When calculating a model value that depends on the $\bar{\beta}$ values e.g. the baseline hazards, it is possible to either use the $\bar{\beta}$ values obtained per bootstrap or the median $\bar{\beta}$ values. If there is a correlation between the $\bar{\beta}$ values and the calculated values, the confidence interval related to the individual $\bar{\beta}$ values will normally be larger since that will include the $\bar{\beta}$ related uncertainty. However, when applying the final model, a fixed set of $\bar{\beta}$ values will be used, thus in comparing results between the centres, it is preferable to use uncertainties that do not include the $\bar{\beta}$ uncertainty, since that will be the same in all the centres. In the following, the reported confidence intervals are therefore based on calculations using median $\bar{\beta}$ values, but both sets of confidence intervals are calculated in the supplied Matlab code. As expected, the baseline functions, seen in figure 4, are not equal between the institutions since different baseline functions were used to create the simulation data (since the baseline values are directly related to the linear predictor, the baseline values in figure 4 should not match the simulated values since not all seven variables were selected for the final model, but the ordering is as expected). Such baseline differences indicate survival differences between the patient cohorts in the different centres beyond the potential variation within the parameters included in the Cox model.

Before "accepting" a model, it is important to check its performance; does the model actually predict the original data? The following model performance figures, which largely agree with those suggested by Royston et al. [18], are those that the current paper recommends to report when evaluation a Cox model. All the performance plots are part of the Matlab code provided together with this paper.

The first plot (Figure 5) is the cumulative distribution of the linear predictor ($\bar{X}_i \bar{\beta}$) within each centre. Cumulative distributions are used instead of differential distributions due to the patient aggregation done at the local computers. The aggregation makes it impossible to make a decision at the optimisation computer level about the cut-points for a differential distribution, which would be needed to make the centre specific distributions comparable (same bin sizes); however, cumulative distributions are comparable even if the distance between consecutive data points are not the same for all centres. The distribution in figure 5 does not show the performance of the model, but differences in the linear predictor distribution between centres will reflect differences in the spread of predicted survival times among the centres. Thus, differences in the linear predictor show heterogeneity between the patient populations at the different centres. As for the cumulative hazard function, the reported cumulative distribution is aggregated at the local centre before submission to the optimisation computer. In figure 5, all three distributions are very similar, reflecting that the parameters of the simulation data all were sampled from the same clinical distribution.

A simple measure of the model performance is the C-Harrell index. The index calculates the fraction of all patient pair permutations in the sample for which the patient living the longest could have been inferred from the Cox model (in case of censoring, only pairs for which it can be determined which patient is living the longest are included) [19]. The C-Harrell values are calculated within each institution for all bootstraps, and the median values and confidence interval is reported to the optimisation computer. The index can be calculated based on the patients within each bootstrap (in-bag patients) or the patients that were not part of a given bootstrap (out-of-bag patients). Since both values are based on a booststrap approach the two median values will likely be close to each other, but the out-of-bag patient value might have a somewhat larger uncertainty, as typical for cross validated values. The C- Harrell index for the simulated data are shown in figure 6.

Ideally, the Cox model should predict the data for the entire range of linear predictors shown in figure 5. A method to validate that is to test the prediction capability in subcohorts at the individual centres. A possible division could be three sub-cohorts defined by the 25% and 75% quantiles of the linear predictors. The cutpoints for the individual centres are defined at the optimisation computer and can either be set identical for all centres or individualised if there are large differences within the linear predictor distributions shown in figure 5. Figure 7 shows such a division into three subcohorts within each system. For each subcohort, the Kaplan-Meier estimator and related 95% confidence interval is shown together with the Cox expected values for the subcohort. Within the uncertainty intervals, it is seen that the expected and observed curves are close to each other for all subgroups and centres.

Figure 7 helps to provide an overview that the model seems acceptable for all survival time-points (the x-axis), but the sensitivity in terms of the linear predictor is more limited since each subcohort consists of quite a variety of linear predictor values. As a supplement to figure 7 it is thus helpful to show the variation in linear predictor in a more detailed way, but for fixed survival time-points. Figure 8 shows such a model calibration for two time-points for each centre. The time-point and number of groups within the plot is supplied from the optimisation computer. Within each centre and for each bootstrap, the Cox model predicted survival probability is calculated for all patients. Based on the survival probabilities from all bootstraps, percentile cutpoints are calculated such that there is the same number of values within each group. These cutpoints are then used within each bootstrap to define subgroups. Within each subgroup, the Kaplan-Meier estimator and the Cox predictor at the given time-point are calculated. Figure 8 show the median of the Cox predictions on the x-axis and the observed survival on the y-axis. It is seen that the points are positioned close to the line of identity, showing that the model at fixed time-points, but with large variation in linear predictors, does predict the observed data.

**Discussion and Conclusions**

Distributed learning is still not used as extensively as data pooling when models based on more than one centre are created. However, with the ever-increasing awareness of General Data Protection Regulations (GDPR), and other issues across different jurisdictions and countries, there is an increasing interest in methods that can protect patient sensitive data, both for clinical use and for research projects. Thus, distributed learning is expected to be extensively used in the future since there is no need for data pooling in national or international databases. However, any information that is provided from a system always has the potential to leak some information, e.g. a login system stating that the password is wrong has released the information that the correct password is not the one entered. Thus, the problem is not data leakage as such, but the amount of data leakage. This work demonstrates that a Cox optimisation performed within a distributed learning environment might lead to full leakage of all the patient information used during model optimisation. The current article has provided recommendations for how to create a Cox model in a distributed setting and how to validate the performance with minimal data leakage. Since missing data is very common within medical data, the suggested framework has been designed such that data imputation is easy to perform as part of the optimisation.

The take-home message is that what likely seems safe to protect data always needs additional evaluation. Another recent example is within AI deep learning image segmentation. In some cases, based on the segmentation model only, it is possible to reconstruct the 3D images used for model creation to an extent where the image information is sufficient to determine the patient [20]. Thus new technologies partly used to protect patient sensitive data likely need detailed evaluation before large scale commercial use of the technology.

**Conflict of interest**

The Authors declares that there is no conflict of interest

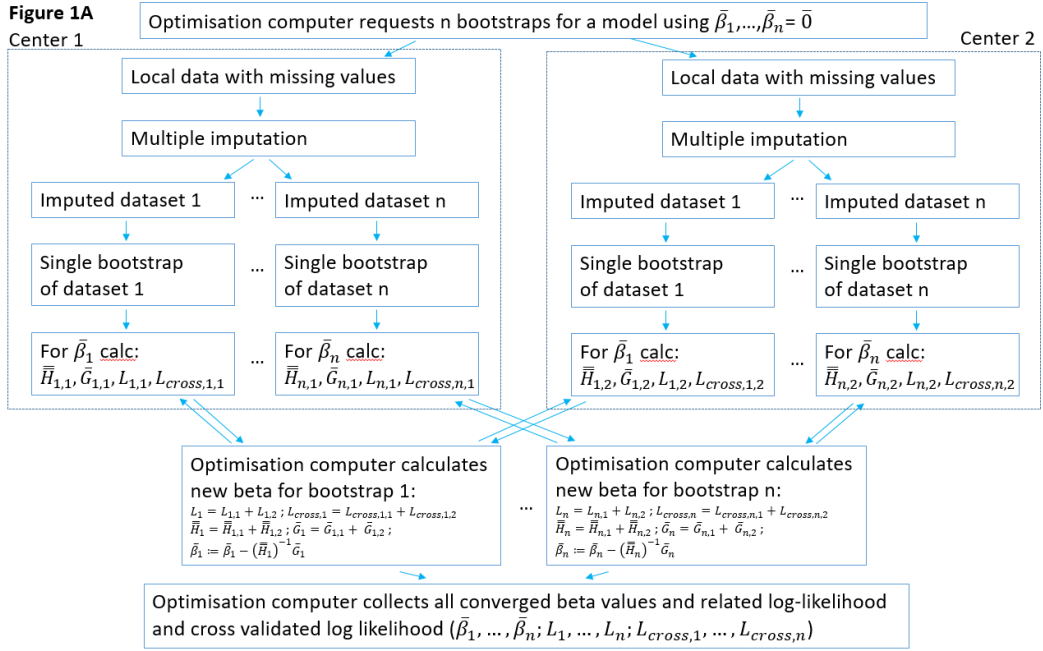

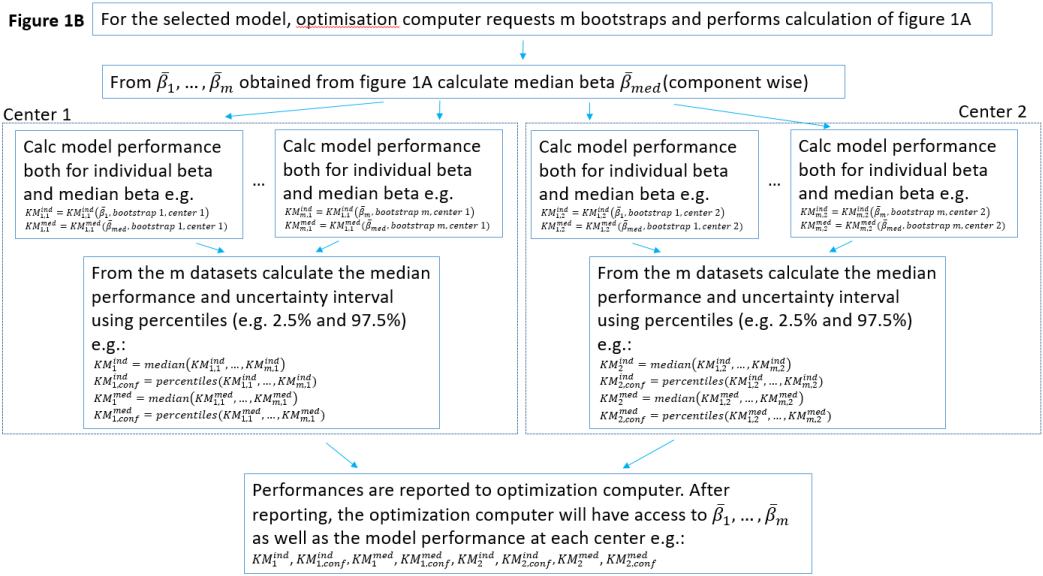

Figure 1: Overview of the calculation process, exemplified for two contributing centres. Part A is related to model selection, while part B is for the model performance. The figure is explained in details in the overview section of the text.

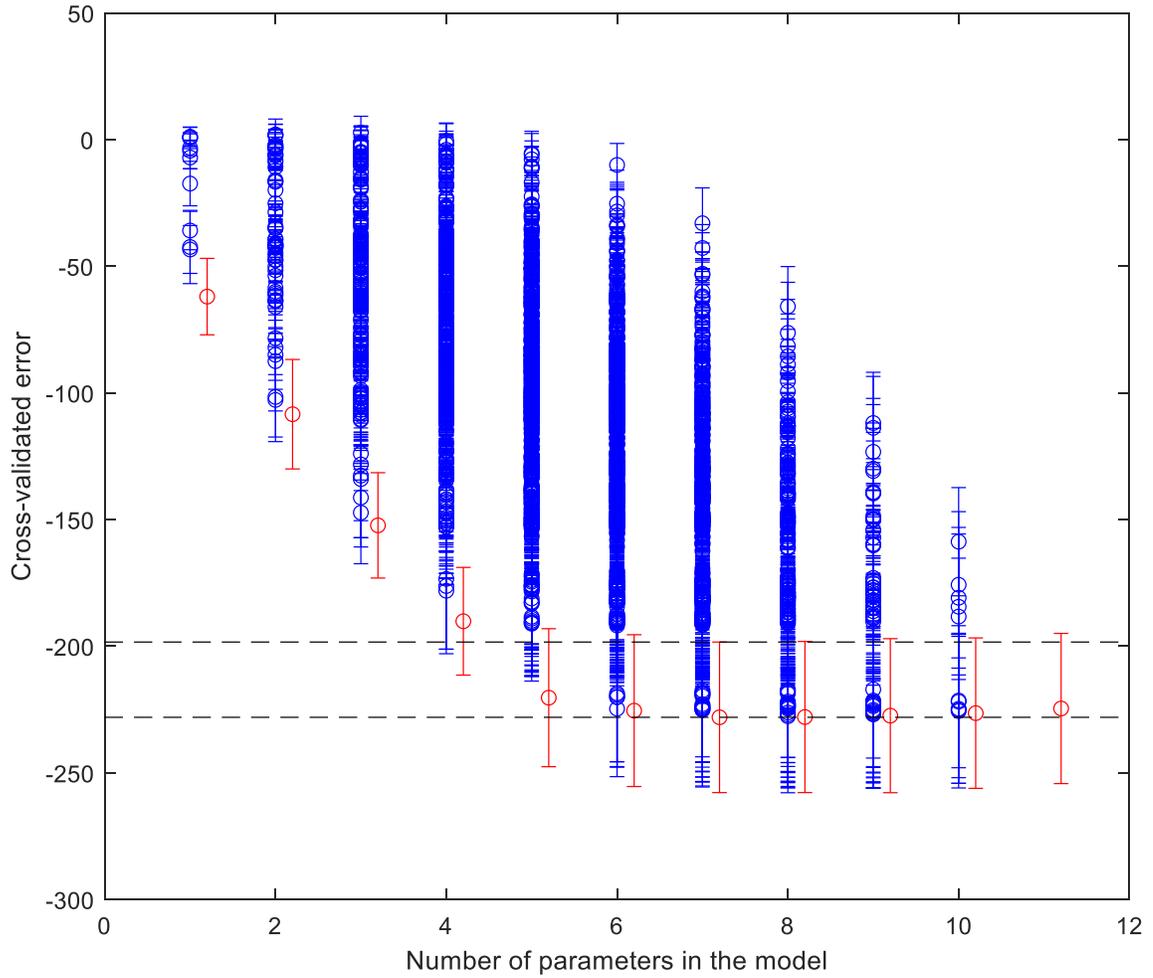

Figure 2: The negative log-likelihoods (corrected for the likelihood of the null model) for each of the 2047 possible models, plotted as a function of the number of parameters within each model. For clarity, the model performing best for a given number of parameters is plotted in red and offset by 0.2 units on the x-axis. The lower dashed curve is at the level of minus log-likelihood of the model performing the best. The upper dashed curve is the upper range of one standard deviation of the best performing model. The best performing model is a model with seven parameters, while the one chosen due to the one-standard-error rule is a model with five parameters.

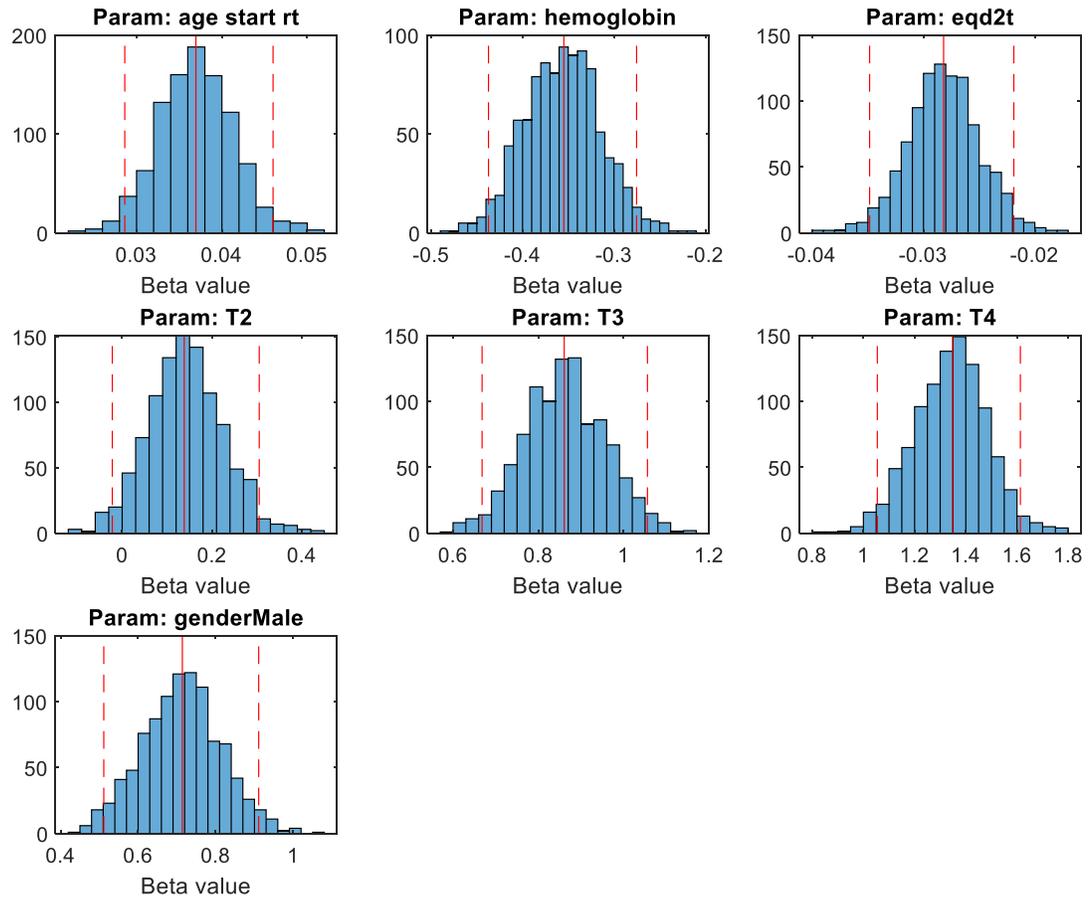

Figure 3: Distribution of the beta values from the 1000 bootstraps performed for the model selected by the one-standard-error rule. The median value (indicated by a solid red line) of the distribution will be the estimated beta value, while the 95% central values will be reported as the confidence interval (indicated by dashed red lines)

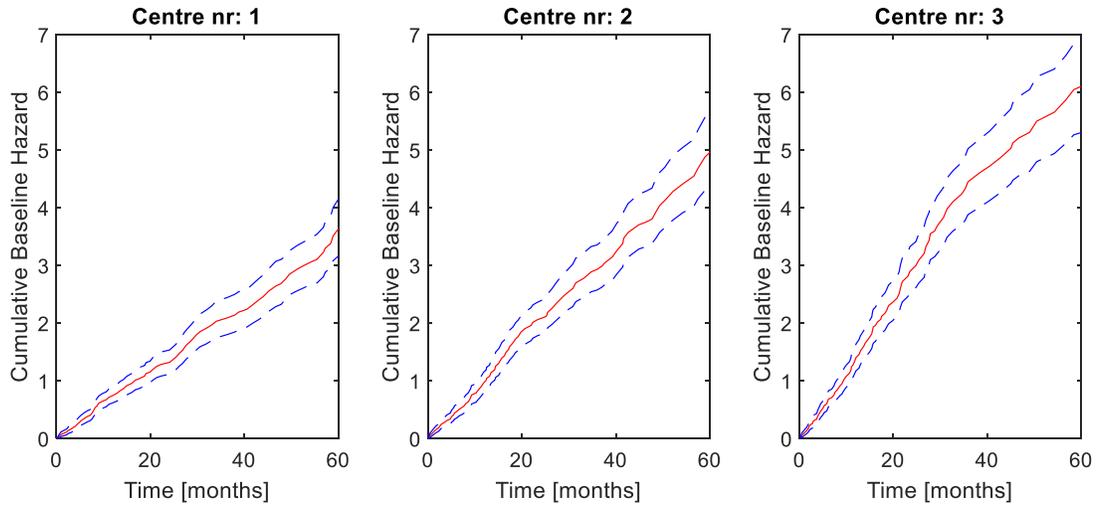

Figure 4: Baseline hazard functions for each of the three institutions. The confidence interval is the 95% confidence interval obtained from the different bootstrapped baseline values.

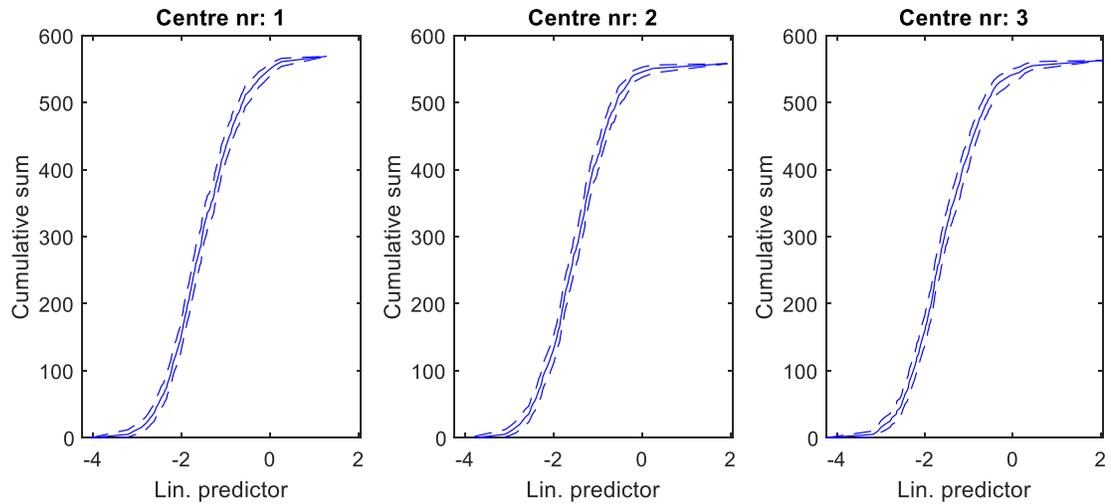

Figure 5: The cumulative distribution of the linear predictor ($\bar{X}_i\bar{\beta}$) within each institution (also known as the prognostic index). A broad distribution indicates a large variation within the model variables at the institution; and thus a larger spread in predicted survival. The confidence interval is the 95% confidence interval based on all the bootstrap predicted distribution values.

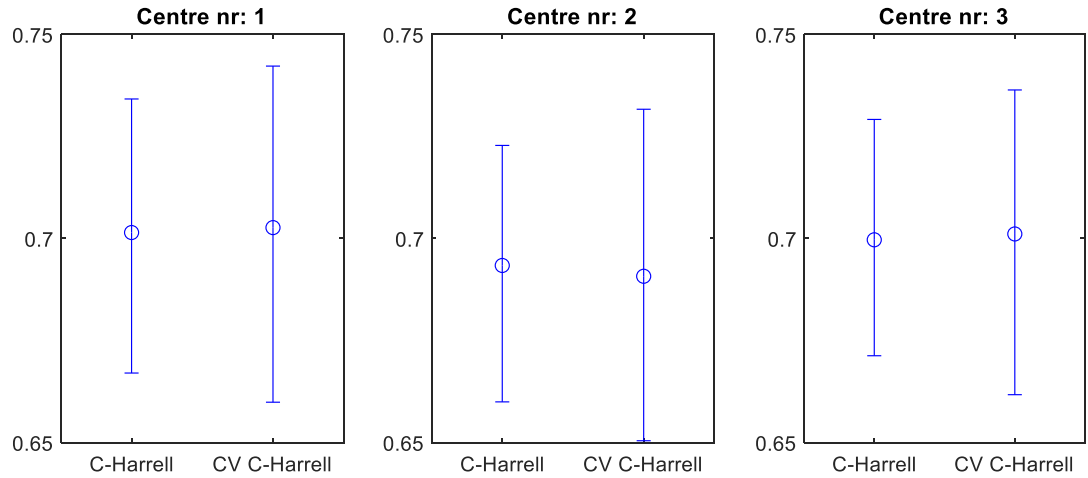

Figure 6: C-Harrell index (Harrell's C-index, concordance index) calculated for each of the three centres. The left column (C-Harrell) in each plot is based on the in-bag bootstrap data, while the right column (CV C-Harrell) is based on the out-of-bag bootstrap data used for cross-validation.

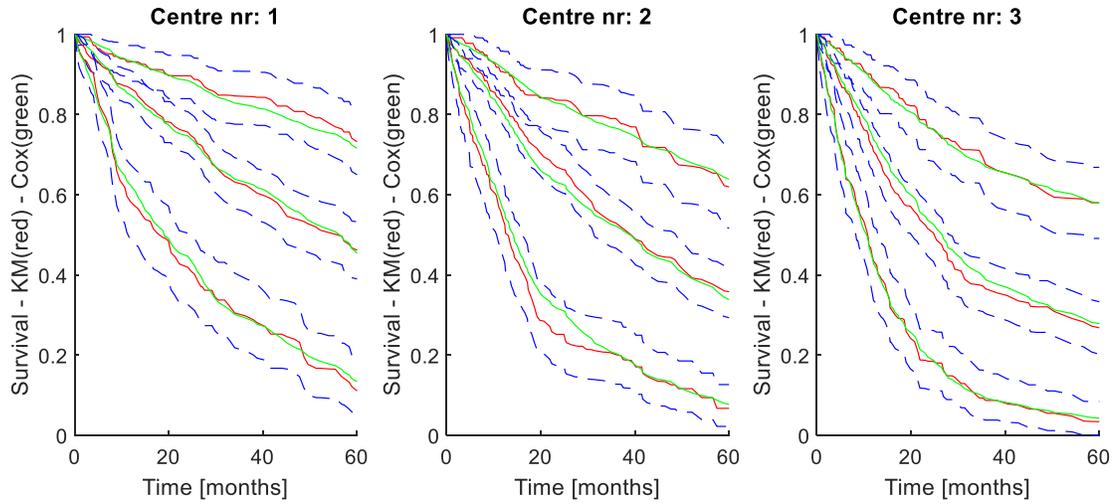

Figure 7: Division of the cohort in subgroups based on the linear predictor. For each subgroup, the red line is the Kaplan-Meier survival estimator, while the green line is the Cox survival model for the given subgroup. The dashed blue lines are the 95% confidence limit of the Kaplan Meier estimator based on the bootstrapped values.

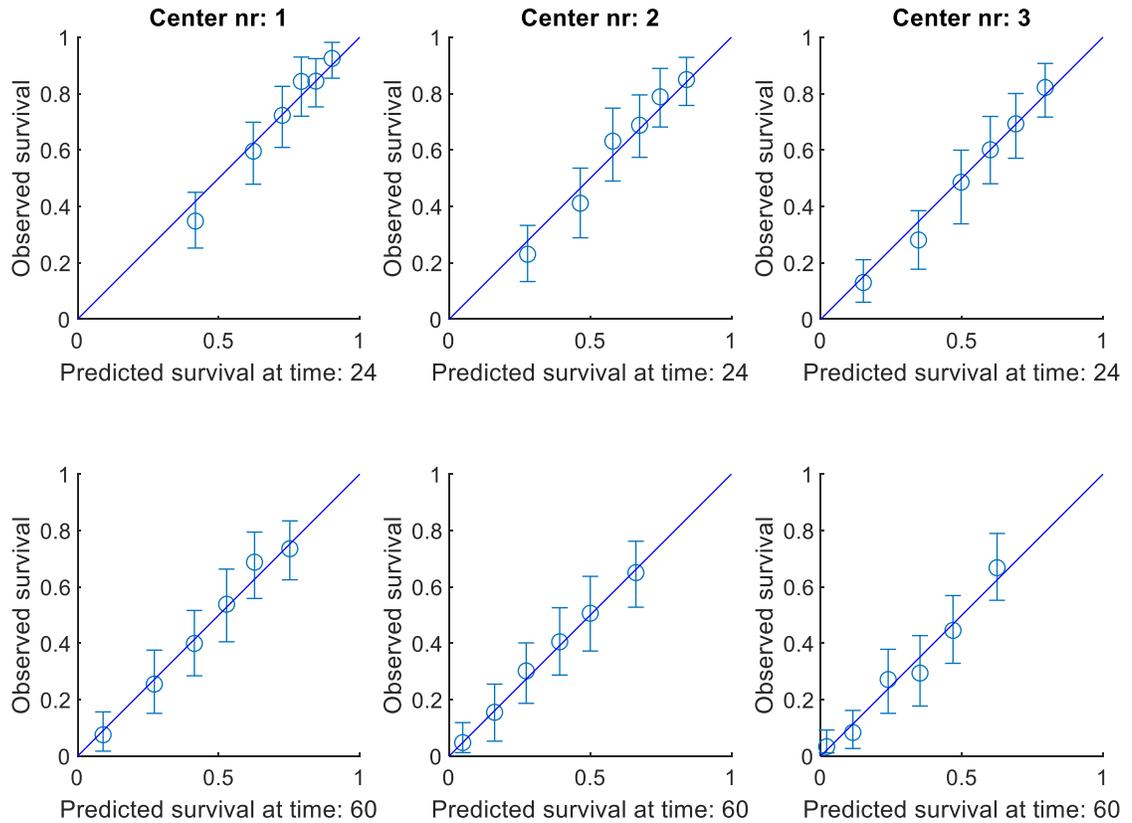

Figure 8: Model calibration plots, based on all the predicted survival probabilities, for all patients and all bootstraps, at the given time-points (here 24 and 60 months – upper and lower row). Cutpoints in survival probability are defined such that survival groups of equal size are created. The predicted survival on the x-axis is given as the median of all survival probabilities per group. The observed survival, on the y-axis, is calculated for each bootstrap and group as the Kaplan-Meier estimator; the plotted value is the bootstrap

median value, while the confidence interval is the 95% confidence interval based on the bootstrap values. The solid line is the identity line (y=x) to guide the eye.

## Appendix 1

In this appendix, the likelihood and derivatives for the Cox model are defined. Also, the iterative approach to estimate the regression parameter for the next iteration is described at the end.

The likelihood for a Cox regression in equation 3 is for the case of no ties (all unique time events, i.e. all $Y_i$ in equation 3 are unique). If ties are present, the likelihood can be specified within the Efron approximation in the following way [21]:

$$L(\bar{\beta}) = \prod_j \frac{\prod_{i \in H_j} \theta_i}{\prod_{l=0}^{m_j-1}\left[\left(\sum_{i:Y_i \geq t_j} \theta_i\right) - \left(\frac{l}{m_j} \sum_{i \in H_j} \theta_i\right)\right]} \qquad Eq.\ A1.\ 1$$

where $m_j = |H_j|$ ($m_j$ is the number of elements in the set $H_j$) with,

$$H_j = \{i|\ Y_i = Y_j\ \wedge\ C_i = 1\} \qquad Eq.\ A1.\ 2$$

So the $H_j$ is the set of index' for which the related Y values are event times and equal to $Y_j$; and $m_j$ is the number of events at the time $Y_j$. If the likelihood is stated as the log-likelihood, it can be written as:

$$l(\bar{\beta}) = \log\left(L(\bar{\beta})\right) \qquad \text{Eq. A1. 3}$$

$$= \sum_j \left[\left(\sum_{i \in H_j} \bar{X}_i \bar{\beta}\right) - \sum_{l=0}^{m_j-1} \log\left(\left(\sum_{i:Y_i \geq t_j} \theta_i\right) - \left(\frac{l}{m_j}\sum_{i \in H_j} \theta_i\right)\right)\right]$$

To create a fast search algorithm for the minimum log-likelihood, it is an advantage to calculate the first and second-order derivatives of the log-likelihood with respect to the individual beta values. We will start by calculating the first derivatives of parts of $l(\bar{\beta})$:

$$\frac{\partial}{\partial \beta_p} \sum_{i \in H_j} \bar{X}_i \bar{\beta} = \sum_{i \in H_j} X_{i,p} \qquad \text{Eq. A1. 4}$$

$$\frac{\partial}{\partial \beta_p} \sum_{i:Y_i \geq t_j} \theta_i = \frac{\partial}{\partial \beta_p} \sum_{i:Y_i \geq t_j} \exp(\bar{X}_i \bar{\beta}) = \sum_{i:Y_i \geq t_j} X_{i,p} \exp(\bar{X}_i \bar{\beta}) \qquad \text{Eq. A1. 5}$$

$$= \sum_{i:Y_i \geq t_j} X_{i,p} \theta_i$$

$$\frac{\partial}{\partial \beta_p} \frac{l}{m_j} \sum_{i \in H_j} \theta_i = \frac{l}{m_j} \sum_{i \in H_j} X_{i,p} \theta_i \qquad \text{Eq. A1. 6}$$

$$\frac{\partial}{\partial \beta_p} \log\left(\left(\sum_{i:Y_i \geq t_j} \theta_i\right) - \left(\frac{l}{m_j} \sum_{i \in H_j} \theta_i\right)\right) \qquad \text{Eq. A1.7}$$

$$= \frac{\left(\sum_{i:Y_i \geq t_j} X_{i,p} \theta_i\right) - \left(\frac{l}{m_j} \sum_{i \in H_j} X_{i,p} \theta_i\right)}{\left(\sum_{i:Y_i \geq t_j} \theta_i\right) - \left(\frac{l}{m_j} \sum_{i \in H_j} \theta_i\right)}$$

Using these results, it is possible to calculate the first derivative of the log-likelihood:

$$\frac{\partial}{\partial \beta_p} l(\bar{\beta}) = \sum_j \left[\left(\sum_{i \in H_j} X_{i,p}\right) - \sum_{l=0}^{m_j-1} \frac{\left(\sum_{i:Y_i \geq t_j} X_{i,p} \theta_i\right) - \left(\frac{l}{m_j} \sum_{i \in H_j} X_{i,p} \theta_i\right)}{\left(\sum_{i:Y_i \geq t_j} \theta_i\right) - \left(\frac{l}{m_j} \sum_{i \in H_j} \theta_i\right)}\right] \qquad \text{Eq. A1.8}$$

Before calculating the second derivative of the likelihood, we will calculate the derivatives of parts of equation A1.8:

$$\frac{\partial}{\partial \beta_q} \sum_{i \in H_j} X_{i,p} = 0 \qquad \text{Eq. A1.9}$$

$$\frac{\partial}{\partial \beta_q} \sum_{i:Y_i \geq t_j} X_{i,p} \theta_i \qquad \text{Eq. A1.10}$$

$$= \frac{\partial}{\partial \beta_q} \sum_{i:Y_i \geq t_j} X_{i,p} \exp(\bar{X}_i \bar{\beta}) = \sum_{i:Y_i \geq t_j} X_{i,p} X_{i,q} \exp(\bar{X}_i \bar{\beta})$$

$$= \sum_{i:Y_i \geq t_j} X_{i,p} X_{i,q} \theta_i$$

$$\frac{\partial}{\partial \beta_q} \frac{l}{m_j} \sum_{i \in H_j} X_{i,p} \theta_i = \frac{l}{m_j} \sum_{i \in H_j} X_{i,p} X_{i,q} \theta_i \qquad \text{Eq. A1.11}$$

$$\frac{\partial}{\partial \beta_q} \sum_{i:Y_i \geq t_j} \theta_i = \sum_{i:Y_i \geq t_j} X_{i,q} \theta_i \qquad \text{Eq. A1.12}$$

$$\frac{\partial}{\partial \beta_q} \frac{l}{m_j} \sum_{i \in H_j} \theta_i = \frac{l}{m_j} \sum_{i \in H_j} X_{i,q} \theta_i \qquad \text{Eq. A1.13}$$

To obtain a compact description of the likelihood and derivatives, it is convenient to define the following quantity and calculate its derivatives:

$$\phi_{j,l,m_j} = \left(\sum_{i:Y_i \geq t_j} \theta_i\right) - \left(\frac{l}{m_j}\sum_{i \in H_j} \theta_i\right) \qquad \text{Eq. A1.14}$$

$$\frac{\partial}{\partial \beta_q}\phi_{j,l,m_j} = \left(\sum_{i:Y_i \geq t_j} X_{i,q}\theta_i\right) - \left(\frac{l}{m_j}\sum_{i \in H_j} X_{i,q}\theta_i\right) \qquad \text{Eq. A1.15}$$

$$\frac{\partial^2 \phi_{j,l,m_j}}{\partial \beta_p \partial \beta_q} = \left(\sum_{i:Y_i \geq t_j} X_{i,p}X_{i,q}\theta_i\right) - \left(\frac{l}{m_j}\sum_{i \in H_j} X_{i,p}X_{i,q}\theta_i\right) \qquad \text{Eq. A1.16}$$

Using equation A1.9-A1.16 it is possible to calculate the second-order derivative of the log-likelihood

$$\frac{\partial}{\partial \beta_q \partial \beta_p} l(\bar{\beta})$$ 
*Eq. A1.17*

$$= \frac{\partial}{\partial \beta_q} \sum_j \left[ \left( \sum_{i \in H_j} X_{i,p} \right) - \sum_{l=0}^{m_j-1} \frac{\left( \sum_{i:Y_i \geq t_j} X_{i,p} \theta_i \right) - \left( \frac{l}{m_j} \sum_{i \in H_j} X_{i,p} \theta_i \right)}{\left( \sum_{i:Y_i \geq t_j} \theta_i \right) - \left( \frac{l}{m_j} \sum_{i \in H_j} \theta_i \right)} \right]$$

$$= -\sum_j \sum_{l=0}^{m_j-1} \frac{\partial}{\partial \beta_q} \frac{\left( \sum_{i:Y_i \geq t_j} X_{i,p} \theta_i \right) - \left( \frac{l}{m_j} \sum_{i \in H_j} X_{i,p} \theta_i \right)}{\left( \sum_{i:Y_i \geq t_j} \theta_i \right) - \left( \frac{l}{m_j} \sum_{i \in H_j} \theta_i \right)}$$

$$= -\sum_j \sum_{l=0}^{m_j-1} \frac{\partial}{\partial \beta_q} \frac{\frac{\partial}{\partial \beta_p} \phi_{j,l,m_j}}{\phi_{j,l,m_j}}$$

$$= -\sum_j \sum_{l=0}^{m_j-1} \frac{\phi_{j,l,m_j} \frac{\partial^2 \phi_{j,l,m_j}}{\partial \beta_q \partial \beta_p} - \frac{\partial \phi_{j,l,m_j}}{\partial \beta_p} \frac{\partial \phi_{j,l,m_j}}{\partial \beta_q}}{\phi_{j,l,m_j}^2}$$

$$= -\sum_j \sum_{l=0}^{m_j-1} \frac{\frac{\partial^2 \phi_{j,l,m_j}}{\partial \beta_q \partial \beta_p}}{\phi_{j,l,m_j}} - \frac{\frac{\partial \phi_{j,l,m_j}}{\partial \beta_p} \frac{\partial \phi_{j,l,m_j}}{\partial \beta_q}}{\phi_{j,l,m_j}^2}$$

Using the $\phi_{j,l,m_j}$ notation the log-likelihood and first and second-order derivatives can thus be stated as follows:

$$l(\bar{\beta}) = \sum_j \left[ \left( \sum_{i \in H_j} \bar{X}_i \bar{\beta} \right) - \left( \sum_{l=0}^{m_j-1} \log(\phi_{j,l,m_j}) \right) \right]$$
*Eq. A1.18*

$$\frac{\partial}{\partial \beta_p} l(\bar{\beta}) = \sum_j \left[ \left( \sum_{i \in H_j} X_{i,p} \right) - \sum_{l=0}^{m_j-1} \frac{\frac{\partial}{\partial \beta_p} \phi_{j,l,m_j}}{\phi_{j,l,m_j}} \right] \qquad \text{Eq. A1. 19}$$

$$\frac{\partial}{\partial \beta_q \partial \beta_p} l(\bar{\beta}) = -\sum_j \sum_{l=0}^{m_j-1} \frac{\frac{\partial^2 \phi_{j,l,m_j}}{\partial \beta_q \partial \beta_p}}{\phi_{j,l,m_j}} - \frac{\frac{\partial \phi_{j,l,m}}{\partial \beta_p} \frac{\partial \phi_{j,l,m_j}}{\partial \beta_q}}{\phi_{j,l,m_j}^2} \qquad \text{Eq. A1. 20}$$

$$\phi_{j,l,m_j} = \left( \sum_{i: Y_i \geq t_j} \theta_i \right) - \left( \frac{l}{m_j} \sum_{i \in H_j} \theta_i \right) \qquad \text{Eq. A1. 21}$$

$$\frac{\partial}{\partial \beta_q} \phi_{j,l,m_j} = \left( \sum_{i: Y_i \geq t_j} X_{i,q} \theta_i \right) - \left( \frac{l}{m_j} \sum_{i \in H_j} X_{i,q} \theta_i \right) \qquad \text{Eq. A1. 22}$$

$$\frac{\partial^2 \phi_{j,l,m_j}}{\partial \beta_p \partial \beta_q} = \left( \sum_{i: Y_i \geq t_j} X_{i,p} X_{i,q} \theta_i \right) - \left( \frac{l}{m_j} \sum_{i \in H_j} X_{i,p} X_{i,q} \theta_i \right) \qquad \text{Eq. A1. 23}$$

The equations A1.18 to A1.23 are used in the Matlab program "PartialLikelihoodCox", using variable names that reflect the variable naming in the above equations.

The single and double derivatives are used to obtain the next estimate of $\bar{\beta}$ in the iterative process. The process is standard practice, but for completeness is described below.

For a given value of $\bar{\beta}$ named $\bar{\beta}_0$, the gradient of the log-likelihood can be approximated by a Taylor expansion for each component of $\bar{\beta}$. Thus we can state:

$$\frac{\partial}{\partial \beta_p} l(\bar{\beta}) = \frac{\partial}{\partial \beta_p} l(\bar{\beta})|_{\bar{\beta}=\bar{\beta}_0} + \sum_q \frac{\partial}{\partial \beta_q \partial \beta_p} l(\bar{\beta})|_{\bar{\beta}=\bar{\beta}_0} (\beta_q - \beta_{q,0})$$

Eq. A1.24

which in matrix notation can be stated as:

$$\nabla l(\bar{\beta}) = \nabla l(\bar{\beta})|_{\bar{\beta}=\bar{\beta}_0} + \bar{\bar{H}}|_{\bar{\beta}=\bar{\beta}_0} (\bar{\beta} - \bar{\beta}_0)$$

Eq. A1.25

where $\bar{\bar{H}}$ is the Hessian matrix that in entry p,q was the double derivative of the log-likelihood with respect to $\beta_p$ and $\beta_q$ (right-hand side of A1.20) and $\nabla l(\bar{\beta})$ is the gradient of the log-likelihood (each component of the gradient can be calculated using equation A1.19). Since we are searching for a minimum, the gradient at that point will be a zero vector. Thus equation A1.25 can be used to determine $\bar{\beta} - \bar{\beta}_0$ by setting the left-hand side of equation A1.25 equal to zero. The next value of $\bar{\beta}$ can thus be obtained as:

$$\bar{\beta} - \bar{\beta}_0 = -(\bar{\bar{H}}|_{\bar{\beta}=\bar{\beta}_0})^{-1} \nabla l(\bar{\beta})|_{\bar{\beta}=\bar{\beta}_0}$$

Eq. A1.26

Since equation A1.25 is only an approximation to first-order, equation A1.26 needs to be iterated to convergence, which often is achieved within 12-20 iterations.

**Appendix 2**

The supplied Matlab code is documented with comments within it to increase the readability. However, an overview of the structure and input data is given below, aiming to make it simpler to understand. The code is divided into two main parts CentralComputer and LocalComputer. The code within CentralComputer is for the OptimisationComputer, while LocalComputer is a class definition that is used to simulate the local computer with access to the patient sensitive data. Besides these two files, a number of others are provided in the src directory utilised by the two main files.

*Overall structure of CentralComputer:*

The main code is only about 90 lines and can be divided into the following sections

1) Initialisation of parameters used during optimisation
2) Initialisation of the local computers – create instances of the LocalComputer Class
3) Optimisation of all the models included in the cross-validation
4) Selection of the final model
5) Optimisation of the final model
6) Plotting of the results from the final model

*Overall structure of LocalComputer:*

LocalComputer is a class definition that can simulate the behaviour of a local computer. The code is about 60 lines, and LocalComputer has only three functions (methods) that OptimisationComputer can use to communicate with the local computer:

1) Instruct LocalComputer to create n sets of imputed data, each followed by one bootstrap (to create the data sets seen in figure 1). Both the in-bag and out-of-bag data will now be available locally. The instruction is only initialisation of LocalComputer, and no data are returned.
2) Calculate the partial log-likelihood for a number of models for each bootstrap created at step 1 and return these (including gradient and Hessian)
3) Return the model performance for the final model based on the local data.

In the final setup, the three methods will be supplied through a web-service. Thus, the only communication between OptimisationComputer and the local computer is through these three methods.

*Parameters for CentralComputer:*

All parameters for the CentralComputer program are defined within the variable Init. Some of the most relevant fields in Init are listed below, followed by a short explanation:

FeaturesToOptimizeFrom: The set of variable names from the local data that is used during cross-validated model selection

FeaturesLevelSets: Sets of variables that should be observed as one variable, e.g. dummy levels T2, T3 and T4 of the T stage variable. Thus, sets defined within this variable will be either in- or excluded from the cross-validated models together.

GlobalSeed: A global seed to the local random generators. By changing this seed, new bootstrap values will be created locally.

NbootCV: The number of bootstraps used during cross-validation

NbootModelInfo: The number of bootstraps performed on the selected final model. This number is normally much larger than NbootCV since these bootstraps are used to define confidence intervals. But using such a large number during cross-validation will often not be needed and will be quite time-consuming.

ConvergTolerance: Convergence of the optimisation is determined upon all changes in regression parameters between two iterations being less than this value.

Alpha: Reported confidence intervals will be 100%(1-alpha).

PI_thresholds: The thresholds used to split the patient cohort into subcohorts used to compare the Kaplan-Meier estimator and the Cox prediction (see figure 7). If no thresholds are supplied, the Kaplan-Meier estimator and Cox prediction will be calculated for the entire cohort without splitting into subgroups.

CalPlotTimePoints: The specific time-points for the calibration plots of figure 8.

NrCalPlotTimePoints: The number of groups that are reported for the calibration plot (figure 8).

NAllowedMissing: The maximum number of allowed missing values per patient for inclusion in the analysis.

*Parameters for LocalComputer:*

The parameters defined locally will normally be read from a local config file. But in the current simulation, the parameters are defined within the class definition for all centres. The local information is the following:

DataPaths: Path to local data that in this simulation is assumed to be stored in a CSV file.

LocalSeed: A local seed that is added to the global seed provided by the optimisation computer. This additional seed prevents the optimisation computer from reproducing the bootstrap selection process, which thus only is known within the local centre.

NrPtPerBin: Specify the number of patients that will be grouped together in the report of cumulative distributions needed to illustrate the model performance (figure 4-7).

*Communication*

It is important to note that the bootstrap approach used in this paper would lead to a lot of communication through the internet service if the local computer were instructed to perform the calculation one bootstrap at a time. But the system is created such that the local computers can be requested to calculate the needed information for a complete set of models (variation of included variables for parameter selection) and a prespecified number of bootstraps within one internet service instruction. Thus even though many models are calculated (the number of combinations of variables times the number of bootstraps), the number of internet communications is limited to the number of iterations needed to reach convergence. In the attached program, the local computer reports both the likelihood and the related gradient and Hessian; thus, convergence is typically reached within 10-20 iterations.